\documentclass[sigconf,natbib=false]{acmart}
\renewcommand\footnotetextcopyrightpermission[1]{} 
\usepackage{graphicx}
\usepackage{subcaption}
\usepackage{bbding}
\usepackage{multirow}
\usepackage{hyperref}
\usepackage{makecell}

\acmConference[Work in Progress]{}{WIP}{2024}

\RequirePackage[
  datamodel=acmdatamodel,
  style=acmnumeric,
  ]{biblatex}

\bibliography{sample-base.bib}

\begin{document}

\title{HBot: A Chatbot for Healthcare Applications in Traditional Chinese Medicine Based on Human Body 3D Visualization}

\author{Bolin Zhang}
\affiliation{%
  \institution{Harbin Institute of Technology}
  \city{Harbin}
  \country{China}
}
\email{brolin@hit.edu.cn}

\author{Zhiwei Yi}
\affiliation{%
  \institution{Harbin Institute of Technology}
  \city{Weihai}
  \country{China}
}
\email{22S030149@stu.hit.edu.cn}

\author{Jiahao Wang}
\affiliation{%
  \institution{Chinese Academy of Sciences}
  \city{Beijing}
  \country{China}
}
\email{jiahaowang0917@gmail.com}

\author{Dianbo Sui}
\affiliation{
  \institution{Harbin Institute of Technology}
  \city{Weihai}
  \country{China}
}
\email{suidianbo@hit.edu.cn}

\author{Zhiying Tu}
\authornote{Corresponding Author.}
\affiliation{%
  \institution{Harbin Institute of Technology}
  \city{Weihai}
  \country{China}
}
\email{tzy\_hit@hit.edu.cn}

\author{Dianhui Chu}
\authornotemark[1]
\affiliation{%
  \institution{Harbin Institute of Technology}
  \city{Weihai}
  \country{China}
}
\email{chudh@hit.edu.cn}

\renewcommand{\shortauthors}{Bolin Zhang et al.}

\begin{abstract}

The unique diagnosis and treatment techniques and remarkable clinical efficacy of traditional Chinese medicine (TCM) make it play an important role in the field of elderly care and healthcare, especially in the rehabilitation of some common chronic diseases of the elderly. Therefore, building a TCM chatbot for healthcare application will help users obtain consultation services in a direct and natural way. However, concepts such as acupuncture points (acupoints) and meridians involved in TCM always appear in the consultation, which cannot be displayed intuitively. To this end, we develop a \textbf{h}ealthcare chat\textbf{bot} (HBot) based on a human body model in 3D and knowledge graph, which provides conversational services such as knowledge Q\&A, prescription recommendation, moxibustion therapy recommendation, and acupoint search. When specific acupoints are involved in the conversations between user and HBot, the 3D body will jump to the corresponding acupoints and highlight them. Moreover, Hbot can also be used in training scenarios to accelerate the teaching process of TCM by intuitively displaying acupuncture points and knowledge cards. The demonstration video is available at \url{https://www.youtube.com/watch?v=UhQhutSKkTU}. Our code and dataset are publicly available at Gitee\footnote{\url{https://gitee.com/plabrolin/interactive-3d-acup.git}}.

\end{abstract}

\ccsdesc[500]{Information systems~Computing platforms}
\ccsdesc[500]{Computing methodologies~Natural language generation}

\keywords{Dialog systems, Traditional Chinese Medicine, 3D Visualization}

\maketitle

\section{INTRODUCTION}
Online medical dialogue systems have been playing an increasingly important role\cite{intro1} in healthcare and medical care, with increasing adoption of such chatbots by patients, caregivers, and clinicians. Chatbots support a more direct and natural way of human-computer interaction via text-based or voice-based communication methods, which makes them suitable for a variety of target groups from young children to the elderly\cite{intro3} and is an ideal candidate in the healthcare field for remote interventions on patients after discharge. To this end, the amount of medical dialogue systems is rising, and they cover a wide range of categories include mental health, physical health, health information, patient assistance, physician assistance, cognitive or developmental health\cite{AIDoc}.

However, the TCM-related dialogue system has received less attention. Existing works mainly focuses on TCM diagnosis of diseases\cite{diagnosis}, medical record management\cite{Record} and knowledge base construction\cite{knowledge-base}, while they do not have the ability to communicate with humans in natural language. Since the \href{https://en.wikipedia.org/wiki/Acupuncture}{acupuncture} points and meridians involved in TCM always appear in the consultation, we hope to tell users the clear location of these acupoints in an intuitive way. Therefore, we develop a \textbf{h}ealthcare chat\textbf{bot} (HBot) based on a human body model in 3D and knowledge graph, which provides conversational services such as knowledge Q\&A, prescription recommendation, moxibustion therapy recommendation, and acupoint search. 

\begin{figure*}[!h]
	\centering
	\includegraphics[scale=0.3]{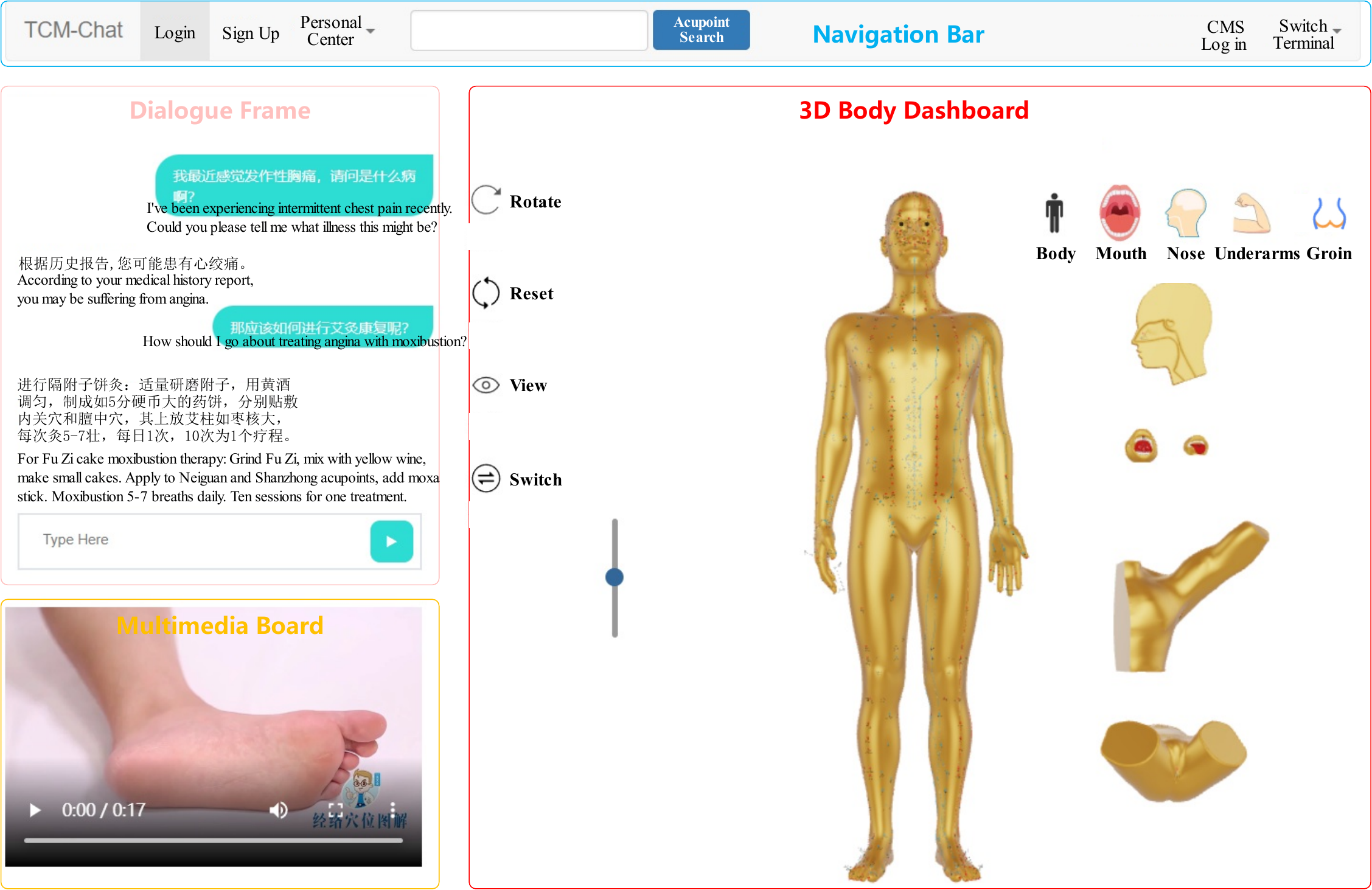}
	\caption{
	The interface of HBot.
	}
	\label{interface}
\end{figure*}

The user interface for this system is shown in Fig~\ref{interface}. The interface is divided into four parts: i)the top area is the navigation bar where users can search an acupoint visually, select registration or login page, and obtain API open capability. ii) the right area is the dashboard of 3D body where users can switch parts of the human body, zoom in\&out and rotate the body, and click on specific acupoints iii) the left-top area is the dialogue frame where users can chat with HBot in text form and give instructions to operate the 3D body, and iv) the left-bottom area is the multimedia board where the pictures or videos related to acupuncture points and meridians will be displayed to help users understand the TCM-knowledge.

The main contributions are summarized as follows:
\begin{itemize}
    \item For the localization and display of acupoints involved in dialogues, we propose an interactive human body in 3D.
    \item To support the consultation services, we construct a knowledge graph (TCM-KG) and annotate a document-level entity\&relation extraction dataset in TCM.
    \item Human evaluation results demonstrate the robustness of Hbot. Functions such as intent detection, entity\&relation extraction, and knowledge query will be publicly released in our platform as open APIs.
\end{itemize}

\section{System Architecture}

\begin{figure}[!h]
	\centering
	\includegraphics[scale=0.42]{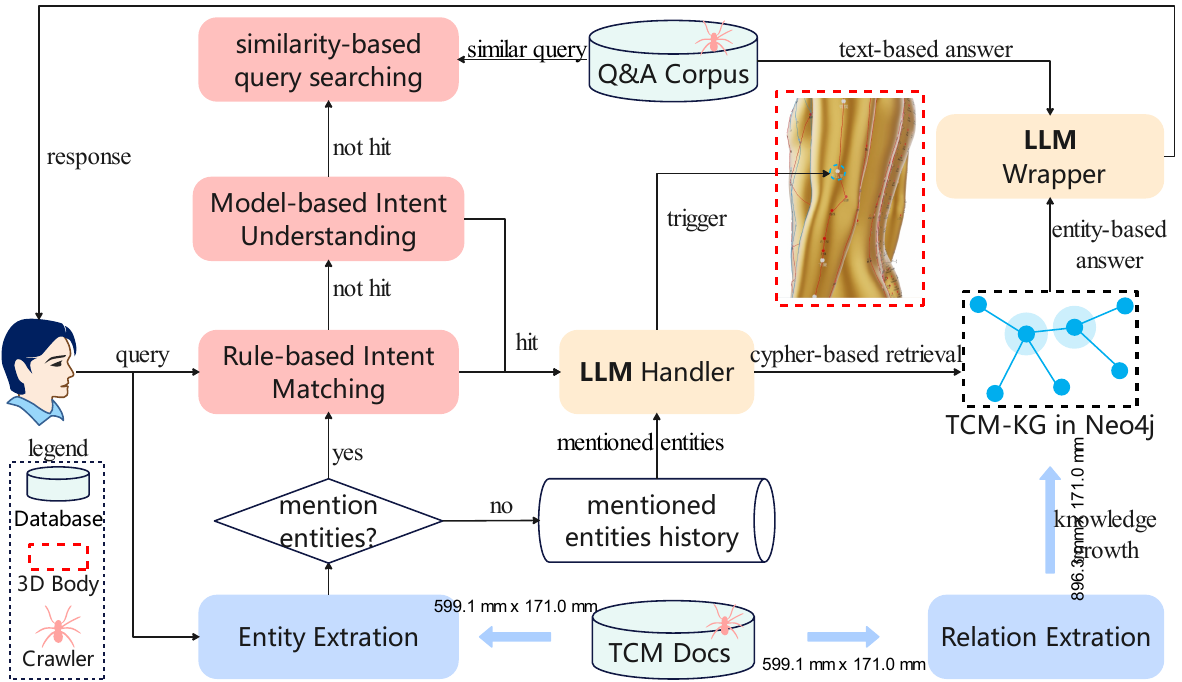}

	\caption{
	The interaction flow of HBot.
	}
	\label{architecture}

\end{figure}

The architecture of the HBot system is shown in Fig~\ref{architecture}, where illustrates an example of interaction flow. HBot contains 6 key modules, Interactive 3D Body, User Intent Detection, Entity\&Relation Extraction, LLM Handler, LLM Wrapper, and Knowledge Graph Construction. User Intent Detection module first classifies user queries and the slot filling task will be finished by the entity extraction module. To achieve multi-turns of dialogue, the history of mentioned entities and user intents will be be saved in a list for dialogue state tracking. According to the dialogue tracking history, the LLM Handler decides whether to trigger the module of Interactive 3D Body or retrieves from the knowledge graph (TCM-KG). Then, the execution result of LLM Handler will be fed into the LLM Wrapper to generate the response. Based on Knowledge Graph Construction, the triples obtained from Entity\&Relation Extraction will be imported into Neo4j for the knowledge growth of TCM-KG. The module of Interactive 3D Body will be activated when user give an instruction or the acupoints are involved in the conversation. These modules are described in detail in the following sections.
 
\section{Interactive 3D Body}
To visually display the meridians and acupoints of the human, we constructed an interactive 3D model of the human body. Considering compatibility of operating systems, simplicity of installation and diversity of interactions, we choose to embed our model into a web page and users can interact with the 3D body only through a web browser. Moreover, we pack the operations of the 3D body into javascript API for the developer to call. The implementation of the 3D body is the following steps:

1.According to the State Standard of the Location of Acupoints\cite{acup}, we manually make an adult male body and mark 409 acupoints by using 3D graphics software (\href{https://www.blender.org/}{blender}).

2.Using the GLTFLoader of \href{https://threejs.org/docs/index.html}{three.js}, the 3D object files (.glb and .hdr) of the body will be loaded in the web page for real-time rendering and displaying.

3.Use the camera controller (controls.js) of three.js to implement 360-degree rotation of the model, switching, zooming and other operations. 

4.Using the event listener functions (onclick, onchange, onmouseover, etc) of Javascript, the switching of body components and the selection and restoration of acupuncture points can be realized. 

\section{User Intent Detection}
User Intent Detection is framed as a sentence classification task that aims to identify intents from human utterances, and plays the key role in dialogue systems\cite{AbroQAA22}. However, training intent detection models relies upon a large set of user utterances paired with intents, which are predefined by the developers. In practice, the user will express new intents that may not be expected by the tested system, referred to as out-of-define (OOD) intents. To alleviate this issue, we propose a configurable intent detection process by applying three-phase: i)rule-based intent matching ii) model-based intent understanding and iii) similarity-based query searching. 

\textbf{Phase I}: an intent-table is maintained in the Redis database, where one predefined intent is mapped to 5 example utterances and the related regular expressions. We defined two types of intents: Body manipulation related (i.e. rotation, zoom in, zoom out) and dialog content related (i.e. asking about treatment, asking about symptoms). When Hbot is running, the user utterance will be matched to a certain intent based on the regular expressions. If the match fails, Hbot will enter the second phase: model-based intent understanding.

\textbf{Phase II}: we applied the complex intent detection model Conco-ERNIE\cite{conco}, which is effective for both multiple intents and implicit intent detection by exploiting co-occurrence patterns between concepts appeared in medical queries and user intents conveyed by these queries. Given a user query, Conco-ERNIE can predict the probability distribution of the utterance on 21 common medical intents. If the probability of each intent is less than the threshold (0.6), Hbot will enter the third phase: similarity-based query searching.

\textbf{Phase III}: We collected a real question-and-answer corpus between patients and doctors (about 100k Q\&A pairs) from medical online forum\footnote{\url{http://club.xywy.com/list_score.htm}}. To ensure the quality of the corpus, we only select the Q\&A pairs where the question has a bounty and the answer is adopted by the patient. To find the approximate answers, we trained a sentence embedding model SBERT\cite{SBERT} with siamese network architecture to calculate the similarity between the input sentence and each query in the corpus. Then, the answer to the query most similar to the input sentence will be returned to the user. Followed by \cite{ESimCSE}, the positive samples of the input question are constructed through word repetition operation. Since entities(i.e. disease, symptoms) appeared in the medical queries have a great influence on sentence semantics, we construct negative samples by randomly replacing the entities of the same or different types.

\section{Entity\&Relation Extraction} 
\label{sec::extraction}

The task of entity extraction (similar to slot filling) is to identify entities in text with their corresponding type, which is a critical component in spoken dialogue systems\cite{sf}. BERT\cite{bert} with a CRF layer have been widely used in named entity recognition and achieved good performance. So we trained a BERT-CRF model to identify all all TCM-related entities in user utterances. 

\begin{figure}[!h]
	\centering
	\includegraphics[scale=0.7]{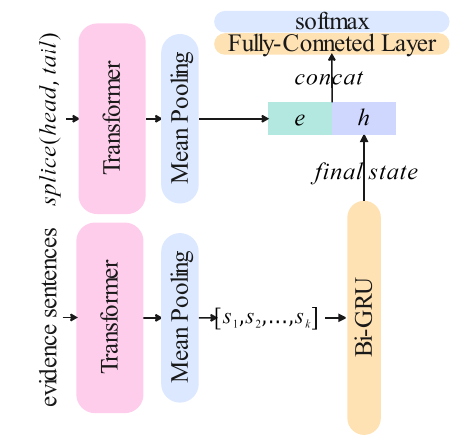}
	\caption{
	The architecture of the RE model.
	}
 \label{REModel}
\end{figure}

The task of relation extraction is to identify the relationship between entities from text\cite{re}, which is essential for the construction of TCM knowledge graph. Under the guidance of medical experts, we finely annotated a small-scale but high-quality document-level entity\&relation extraction dataset (contains 270 documents, 5996 entities, 4685 relations) based on TCM books. Followed by \cite{docred}, we treat relation extraction as a multi-label text classification problem. Given an entity pair (head, tail) and the evidence sentences that used to judge its relation, the spliced text of entity pair will be encoded as $e$ by BERT and the $i$-th evidence sentence will be encoded as $s_i$ by the same BERT. All the representations of evidence sentences $[s_1,s_2,..,s_k]$ will be fed into a Bi-GRU layer, then the final state vector $h$ of Bi-GRU layer and $e$ will be concatenated together and then use a softmax function to compute the probability for each relation type. The relation extraction model is shown as Fig~\ref{REModel} and the performance is shown in Table~\ref{ee-re}.

\begin{table}[!h]
\caption{The performances of different models on the task of entity\&relation extraction}
\label{ee-re}

\begin{center}
    \begin{tabular} {lllll}
    \hline
    \makecell[c]{Task} & \makecell[c]{Model} & \makecell[c]{\textit{precision}} & \makecell[c]{\textit{recall}} & \makecell[c]{\textit{F1}}  \\
    \hline
    \multicolumn{1}{c}{\multirow{3}[2]{*}{\shortstack{Entity Extraction}}} & \makecell[c]{BiGRU+CRF} & 57.52 & 65.99 & 61.47 \\
    & \makecell[c]{\textbf{BERT+CRF}} & 78.30 & 80.86 & 79.47 \\
    & \makecell[c]{ERNIE+CRF} & 76.21 & 79.16 & 77.65 \\
    \hline
    \multicolumn{1}{c}{\multirow{3}[2]{*}{\shortstack{Relation Extraction}}} & \makecell[c]{ERNIE} & 78.46 & 80.67 & 78.72 \\
          & \makecell[c]{\textbf{BERT+GRU}} & 82.36 & 82.72 & 81.56 \\
          & \makecell[c]{ ERNIE+GRU} & 80.15 & 81.07 & 79.63 \\
    \hline
    \end{tabular}
\end{center}

\end{table}

\section{LLM Handler and Wrapper}
Based on the prompt engineering, the LLM Handler utilize the large language model to choose the next action for the given dialogue history. To respond to users' medical questions, the LLM handler retrieves from two types of data sources: documents and knowledge graphs, in order to obtain text-based answers and entity-based answers respectively. The dialogue history, current user intent, and these two types of answers are input into the LLM Wrapper to generate a fluent answer that conforms to the user instructions. Essentially, our LLM wrapper is a RAG (Retrieval-Augmented Generation) system that comprehensively utilizes knowledge from multiple sources and of multiple types. The LLM adapted by both of these modules is ChatGLM3-6B \footnote{\url{https://huggingface.co/THUDM/chatglm3-6b-base}}.

\section{Knowledge Graph Construction}
Knowledge graph is a knowledge base that uses a graph-structured model to integrate data, which is the cornerstone of question answering system. To implement question answering in the domain of TCM healthcare, we construct a TCM-related knowledge graph through three steps: i) ontology modeling ii) instance injection and iii) knowledge growth.

\textbf{Step I}: Followed by ontology development 101 \cite{101}, we create the ontology model in the domain of TCM.  
We first enumerated the important terms in TCM-ontology and defined their properties. Then we defined the relations between different terms.
After that, TCM-ontology contained 12 types of entities, 8 types of relations and 20 types of properties.

\begin{figure}[!h]
	\centering
	\includegraphics[scale=0.125]{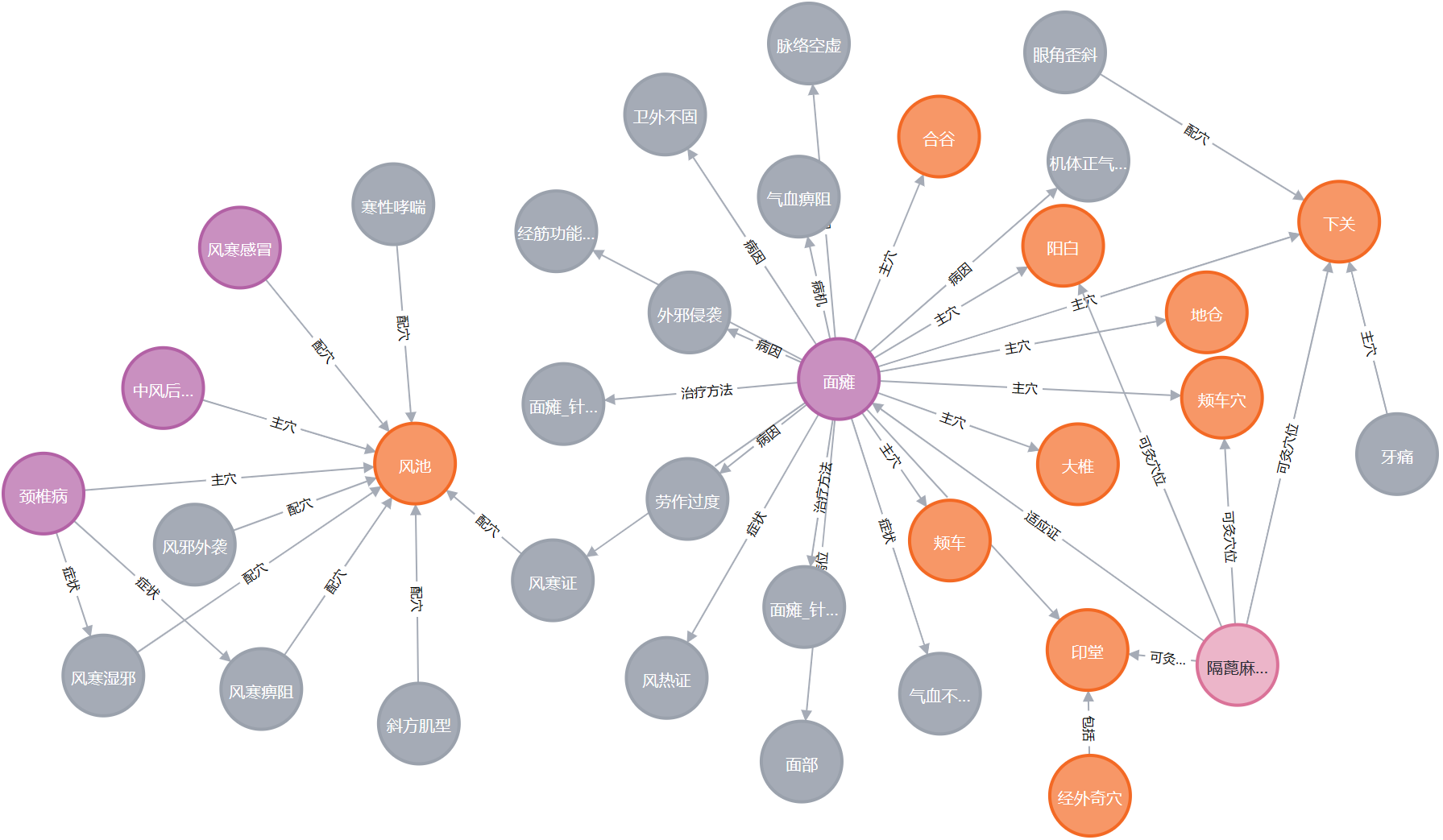}
	\caption{
	A 3-hop subgraph of our TCM-KG stored in Neo4j.
	}
	\label{kg}
\end{figure}

\textbf{Step II}: Since the entities and relations in the extraction dataset mentioned in Sec.~\ref{sec::extraction} are annotated according to TCM-ontology, it is feasible to restored the knowledge triples aligned with the ontology from the dataset directly, and then injected them into the graph database (Neo4j) after deduplication. Through instance injection, we construct a handcrafted TCM domain KG with 37851 triples. Fig~\ref{kg} shows a 3-hop subgraph of with facial paralysis (entity type is Disease) as the central node, where gray nodes represent symptoms, orange nodes represent human acupuncture points, purple nodes represent diseases and pink nodes represent moxibustion therapy. 

\textbf{Step III}: To obtain a larger-scale TCM-KG, we crawled more documents from the National service \href{http://www.gjmlzy.com:83/}{platform} for academic experience of famous TCM doctors, and used the entity\&relation extraction model mentioned in Sec.\ref{sec::extraction} to find more triples from these documents. After manual proofreading, the correct triples are injected into Neo4j. The size of TCM-KG grows from 37851 triples to 48633 triples.

\section{Human Evaluation}
\begin{figure}[!h]
	\centering
	\includegraphics[scale=0.7]{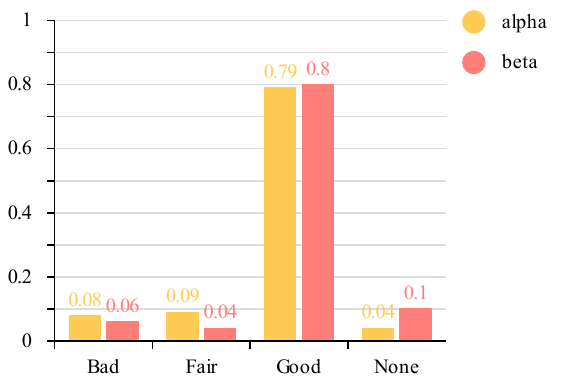}
	\caption{
	The performance of HBot on $\alpha$ and $\beta$ testing.
	}
	\label{abtest}
\end{figure}

Followed by the methodologies of software testing \cite{st}, we conducted \href{https://www.guru99.com/alpha-beta-testing-demystified.html}{$\alpha$\&$\beta$ testing} against user requirements to determine whether HBot satisfies the acceptance criteria. $\alpha$-testing performed to identify all possible issues and bugs before releasing the final product to the end users. A total of 5 members in the development team conducted $\alpha$-testing on HBot from three aspects: the feedback on 3D model operation, the accuracy of user instruction understanding, and the robustness of system at runtime. During $\alpha$-testing, we executed 100 test cases and found 21 bugs, including 10 error feedbacks, 8 responses that did not satisfy instructions, and 3 situations where the system could not run stably. After fixing these bugs, we hired 8 members of our lab to conduct $\beta$-testing on HBot. These people were only informed of the system's features and did not grasp the development details, who can be regarded as end users. During $\beta$-testing, each member was asked to give 15 different instructions to HBot in the form of text in the dialog box. Each instruction is recorded along with the corresponding feedback or response from HBot and the evaluation of user satisfaction. User satisfaction measures whether the feedback or response of HBot meets the expectations of users:
\begin{itemize}
    \item Bad: not at all as expected 
    \item Fair: as expected, with tolerable errors
    \item Good: exactly as expected
    \item None: no feedback or response was returned
\end{itemize}
The performance of HBot on $\alpha$ and $\beta$ testing is shown in Fig~\ref{abtest}. After a round of bug fixes, the responses of HBot are generally more in line with user expectations. However, there are more cases where the user evaluation is none. This is because the testers are developers in $\alpha$-testing and they know where the capability boundaries of the system are, while the testers in $\beta$-testing often give instructions that are outside the scope of the domain (e.g. chat with them, amuse them and ask how much is the surgery). To continuously improve system performance, we developed the evaluation interfaces for end-user and the function of record collection. When the scale of the records is large enough, relevant modules in HBot will be updated offline by analyzing these records.

\section{Conclusions}
To provide TCM-related conversational services such as knowledge Q\&A, prescription recommendation, moxibustion therapy recommendation, and acupoint search, we develop a \textbf{h}ealthcare chat\textbf{bot} (HBot). Based on an interactive human body in 3D, HBot supports the localization and display of acupoints in conversations. Moreover, we annotate an entity\&relation extraction dataset and construct a knowledge graph in TCM to support the slot filling and knowledge retrieval functions of HBot.

\begin{acks}
This work is supported by the National Key R\&D Program of China (2022YFF0903100), the Special Funding Program of Shandong Taishan Scholars Project, the Key projects of Shandong Natural Science Foundation (ZR2020KF019) and Harbin Institute of Technology Graduate Teaching Reform Project (23Z-DZ039).
\end{acks}


\printbibliography

\appendix

\end{document}